\documentclass{article}

\usepackage{arxiv}

\usepackage[utf8]{inputenc} 
\usepackage[T1]{fontenc}    
\usepackage{hyperref}       
\usepackage{url}            
\usepackage{booktabs}       
\usepackage{amsmath}
\usepackage{amsfonts}       
\usepackage{amssymb}
\usepackage{nicefrac}       
\usepackage{microtype}      
\usepackage{cleveref}       
\usepackage{graphicx}
\usepackage{natbib}
\usepackage{doi}
\usepackage{xcolor}         
\usepackage{comment}
\usepackage{dsfont}
\usepackage{subcaption}     
\usepackage{siunitx}

\newcommand{\midb}{\pmb{\mid}}
\mathchardef\mhyphen="2D

\newtheorem{definition}{Definition}
\newtheorem{theorem}{Theorem}
\newtheorem{lemma}{Lemma}
\newtheorem{col}{Corollary}
\newcommand{\bdf}{\begin{definition}\em}
\newcommand{\edf}{\end{definition}}
\newcommand{\bt}{\begin{theorem}\em}
\newcommand{\et}{\end{theorem}}
\newcommand{\bl}{\begin{lemma}\em}
\newcommand{\el}{\end{lemma}}
\newcommand{\bc}{\begin{col}\em}
\newcommand{\ec}{\end{col}}

\newcommand{\bea}{\begin{eqnarray}}
\newcommand{\eea}{\end{eqnarray}}
 
\newcommand{\ben}{\begin{enumerate}}
\newcommand{\een}{\end{enumerate}}

\newcommand{\argmin}{\operatornamewithlimits{argmin}\limits}
\newcommand{\argmax}{\operatornamewithlimits{argmax}\limits}
\renewcommand{\max}{\operatornamewithlimits{max}\limits}

\definecolor{m1}{rgb}{0,0.4470,0.7410}
\definecolor{m2}{rgb}{0.8500,0.3250,0.0980}
\definecolor{m3}{rgb}{0.9290,0.6940,0.1250}
\definecolor{m4}{rgb}{0.4940,0.1840,0.5560}
\definecolor{m5}{rgb}{0.4660,0.6740,0.1880}
\definecolor{m6}{rgb}{0.3010,0.7450,0.9330}
\definecolor{m7}{rgb}{0.6350,0.0780,0.1840}

\newcommand{\vb}[1]{}
\newcommand{\sw}[1]{}

\title{Using Transformers for Teaching and Learning via Symbolic Distillation of Neural A-life Agents}

\usepackage{authblk}

\author[1]{Shuwei Wang}
\author[1]{Vadim Bulitko}
\author[1]{Matthew R.~G.~Brown}
\author[2]{Michael Youngblood}
\author[3]{Ramon Lawrence}
\author[1]{Roscoe Smith}
\author[4]{Shinichi Nakagawa}
\author[5]{William Yeoh}
\affil[1]{Department of Computing Science, University of Alberta, 116 Street and 85 Avenue, Edmonton, AB, T6G 2R3, Canada}
\affil[2]{Filuta AI, 1606 Headway Circle, Austin, TX 78754, USA}
\affil[3]{Department of Computer Science, Data Science, University of British Columbia, 6200 University Blvd, Vancouver, BC Canada V6T 1Z4, Canada}
\affil[4]{Department of Biology, University of Alberta, 116 Street and 85 Avenue, Edmonton, AB, T6G 2R3, Canada}
\affil[5]{McKelvey School of Engineering, Washington University in St.\ Louis, 1 Brookings Drive, St.\ Louis, MO 63130-4899, USA}
\affil[ ]{\texttt{\{shuwei4,bulitko\}@ualberta.ca}, \texttt{mrbrown23@gmail.com}, \texttt{gmyoungblood@gmail.com}, \texttt{ramon.lawrence@ubc.ca}, \texttt{roscoe1@ualberta.ca}, \texttt{snakagaw@ualberta.ca}, \texttt{william.yeoh@gmail.com}}

\renewcommand{\shorttitle}{Transformers for Symbolic Distillation of Neural A-life Agents}

\hypersetup{
pdftitle={Using Transformers for Teaching and Learning via Symbolic Distillation of Neural A-life Agents},
pdfauthor={Shuwei Wang, Vadim Bulitko, Matthew R. G. Brown, Michael Youngblood, Ramon Lawrence, Roscoe Smith, Shinichi Nakagawa, William Yeoh},
pdfkeywords={artificial life, symbolic distillation, transformers, neurosymbolic AI, teaching and learning},
colorlinks=true, linkcolor=blue, citecolor=blue, urlcolor=blue,
}

\begin{document}
\maketitle

\begin{abstract}
Neural networks controlling behavior of A-life agents can evolve over multiple generations. While effective in making the agents survive, such networks are generally opaque to a human observer. Furthermore, the neural weights cannot be transferred between agents with different neural architectures. Distilling such neural networks into compact symbolic policies can help humans understand agent behavior in A-life. Additionally, it allows the agents to share their knowledge in a compact fashion, improving population-level learning.  
Distilling a neural policy into a symbolic policy is typically performed by searching the space of symbolic policies to find one that approximates the neural policy. However, even with a domain-specific language the space of policies can be large, necessitating incomplete stochastic search such as genetic algorithms. Yet the distillation remains computationally expensive.
We propose to instead use a transformer pre-trained on readily available random percept-action data. Doing so speeds up distillation substantially. We derive the transformer architecture via a neural architecture search. Empirical results compare different distillation methods and the influence of the symbolic space on A-life population dynamics through self-reflection and teaching and learning.
\end{abstract}


\section{Introduction}
\label{sec:introduction}

While neural networks have enabled breakthrough performance in many applications over the last decade~\citep{mnih2015human} their learned knowledge is encoded non-transparently via a large number of trained parameters (weights). The training process itself can be unstable and the learned knowledge cannot be compactly transferred between different neural architectures. Neurosymbolic Artificial Intelligence (AI)~\citep{chaudhuri2021neurosymbolic,kautz2022third,10.1093/nsr/nwac035,garcez2023neurosymbolic} seeks to address these three shortcomings by complementing neurally represented knowledge with symbolically represented knowledge. One way of doing so is {\em distillation} of neural control policies: finding a compact symbolic policy of similar behavior~\citep{verma2018programmatically}. 

The distillation methods used in this process often come from the field of symbolic regression~\citep{gulwani2017program} and involve searching a space of programs~\citep{odena2020bustle}. 
Because the space of programs is usually large, incomplete stochastic methods are utilized. In particular, stochastic search in the form of simulated annealing~\citep{medeiros2022can} and genetic algorithms (GA)~\citep{koza1992r, cranmer2020discovering} is common in program synthesis~\citep{bulitko2022portability,wang2023synthesizing} and neural distillation~\citep{deproost2024human}. While it has been successfully used for neurosymbolic AI~\citep{aifeynman, acharya2023neurosymbolic} it can be slow and unreliable. 

In this paper we adopt the idea of symbolic regression via pre-trained transformers~\citep{biggio2021neural, vaswani2017attention} and synthesize compact symbolic policies of neural agents evolved within A-life. The transformer receives a set of an agent's percepts and the actions taken by the agent and produces a series of tokens describing a symbolic policy that approximately matches the percept-action pairs. We use a neural architecture search~\citep{chen2021autoformer} to automatically find an effective neural architecture for the transformer. Once trained the transformer can be enhanced with bagging~\citep{kamienny2022end} and beam search~\citep{vaswani2017attention}. We show that not only can it surpass the accuracy of symbolic policies produced by genetic algorithms but its distillation is also substantially faster.

The distillation allows a population of A-life agents to exchange their knowledge in a compact, symbolic fashion. Such knowledge sharing enables population-level learning. The compact nature of the symbolic policies also makes them human-readable which may lead to theoretical verifiability~\citep{zhu2019inductive,bastani2018verifiable,stevens2023solving} and human insight~\citep{10645663}. The symbolic policies can also be used to stabilize the neural training in an iterative manner~\citep{verma2018programmatically, verma2019imitation} and to communicate knowledge between agents with different neural architectures~\citep{neat}.  

The paper makes the following contributions. We propose a method for automatically designing and training a transformer for distilling neural policies of A-life agents. Once trained the transformer surpasses the distillation accuracy of a genetic algorithm on average while being substantially faster. The short per-policy distillation time makes teaching and learning within A-life more practical. In particular, we consider the impact of the space of symbolic policies used for the distillation on teaching and learning in the population. Finally, we discuss human readability of symbolic policies. Our code and data are available to other researchers upon request.


\section{Problem Formulation}
\label{sec:problemFormulation}

A neural control policy $\pi$ of an A-life agent maps the agent's percept $p$ to an action $\pi(p) = a$. The policy $\pi$ is not directly available to the distiller which instead receives a set of input (agent percept) and output (agent action) pairs: $\left\{(p_i,\pi(p_i))\right\}$. Then given a space $ M $ of symbolic policies the {\em distillation task} for agent policy $\pi$ is to find a policy $m^* \in M$ with the maximum accuracy on the percept-action set $(p_i,a_i)$: $m^* = \argmax_{m \in M} \mathcal{A}(m,\left\{(p_i,\pi(p_i))\right\})$ where $\mathcal{A}$ is the percentage of percepts $p_i$ on which the symbolic policy gives the same action as the agent's neural policy: $m(p_i) = \pi(p_i)$.

We evaluate our distillers on their ability to accurately distill neural agents evolved in A-life. The corresponding performance measure is the {\em distillation accuracy}. We generalize this definition to percept-action pairs whose actions come from other sources (e.g., an unknown symbolic policy). For instance, given a set of symbolic policies one can create a set of percept-action pairs for each of them by running each policy on random percepts. Then the distillation accuracy of a distiller is the proportion of actions produced by a symbolic policy that match the actions of the original policy.

We prefer our distiller to be constructed automatically. We also prefer that it produces a compact symbolic policy to help humans with readability and theoretical verifiability. Within A-life compact symbolic policies are faster to transmit and interpret. Thus ideally the symbolic policies produced by a distiller are more compact than the agent's neural policies that they approximate.

Finally, we consider how symbolic policies produced by a distiller and used for teaching and learning within a population of autonomous agents evolving in A-life affect the population dynamics (e.g., agent life expectancy).


\section{Related Work}
\label{sec:relatedWork}

Discovering a symbolic approximation of an unknown function is related to program synthesis which seeks to find a symbolic program for a given problem specification~\citep{alur2013syntax}. Search-based methods such as random sampling, simulated annealing, genetic algorithms and inductive logic programming have been used~\citep{bulitko2022portability, wang2023synthesizing,10645663,stevens2023solving}. Additionally synthesized programs can be human-explainable and allow for theoretical verifiability~\citep{10645663,stevens2023solving}. Such synthesis methods can also be applied iteratively with synthesized programs added to the domain-specific language as the new building blocks~\citep{hernandez2021speeding}. However, when searching the space of programs program-synthesis methods commonly evaluate each candidate program by running it on a data set. This is computationally expensive. The stochastic nature of the search can also make it unreliable. The latter can be remedied by running a portfolio of such searches and taking the best of several runs~\citep{botea2025} but the computation costs increase even further.

A line of work proposes to aid the combinatorial symbolic search by neural networks~\citep{menon2013machine,ellis2018learning,valkov2018houdini,shah2020learning,ellis2021dreamcoder}. The neural networks are trained on problem specification and program pairs and, in general, a search in the symbolic space is still required which is something we wish to eliminate in solving our problem.

Recent research has applied pre-trained transformers for symbolic regression tasks demonstrating that they can efficiently synthesize symbolic expressions, perform sequence-to-sequence program induction and generalize across a variety of rule-based problems~\citep{biggio2021neural, valipour2021symbolicgpt, kamienny2022end, shojaee2023transformer, lalande2023transformer, vastl2024symformer, wang2026synthesizing}. Analysis reveals that a transformer's internal representations can capture recursive rules and variable binding properties~\citep{brinkmann2024mechanistic,wu2025transformers}. Neurosymbolic paradigms have emerged that combine transformer-based symbolic regressors with evolutionary or preference-based algorithms to enhance robustness when dealing with limited or noisy data~\citep{jha2024evolutionary, lalande2023transformer}. These methods often outperform traditional search-based symbolic program synthesis in terms of efficiency and accuracy. To the best of our knowledge none of these studies has been applied to A-life-style multi-agent systems considered in this paper.

Large language models (LLMs) and transformers have been used in agent-based modeling to control the agents~\citep{jimenez2025multi} and to capture the environment dynamics and agent trajectories in a neural format~\citep{berkovich2025lifegpt,berkovich2025automatagpt,yuan2021agentformer}. None of these tackles our problem of distilling an agent neural policy into a symbolic approximation. 

Work on rule inference trained transformers to produce cellular automata rules given a target phenomenon~\citep{berkovich2025automatagpt}. While our problem is also to produce symbolic policies we seek to have the policies approximate evolved neural policies of individual A-life agents as opposed to global dynamics of the environment. Work on modeling agents from observed trajectories used transformers to capture influences of other agents' policies which can then be used to condition the controlled agent's policy and improve its reinforcement learning~\citep{wallace2025transam}. However, our problem is to generate compact symbolic policies of individual neural agents in A-life.

Finally, our effort in this paper to enable agents to teach and learn from each other is related to the large body of work on explainable AI where AI agents seek to explain their predictions~\citep{DBLP:conf/uic/ChakrabortyTRHA17}, recommendations~\citep{zhang2020explainable}, schedules~\citep{vasileiou23} or plans~\citep{Fox2017ExplainableP,kambhampati2019synthesizing}. When an AI agent is using machine learning (ML) it may be employing ML representations that range from ones that can be analyzed easily (e.g., logistic regression and decision trees) to those that are more complex and can highlight key regions in the input data~\citep{DBLP:journals/corr/SpringenbergDBR14} or features~\citep{DBLP:conf/kdd/ChenCW14, guidotti2018survey, DBLP:conf/uic/ChakrabortyTRHA17} relevant to the solution. Within the context of symbolic planning and sequential decision making the problem is cast as explainable planning~\citep{Fox2017ExplainableP,kambhampati2019synthesizing}. In this subarea a popular approach is one that reconciles the models of the human and AI agents such that they both understand the plan being explained~\citep{chakraborti2017plan}. However, existing work~\citep{chakraborti2017plan,sreedharan2018handling,sreedharan2021using,son2021model,vasileiou2021exploiting} makes simplifying assumptions such as that the AI agent has a complete and accurate mental model of the user and that the explanations are one-shot interactions. Researchers have started to address some of these limitations by generalizing the framework to learn the mental model of users over time~\citep{DBLP:conf/aaai/TangV025} as well as to allow for conversational multi-shot interactions~\citep{DBLP:conf/kr/VasileiouK0ST24}. Our problem can be seen as a combination of both interpretable ML and explainable planning where the neural policy of one agent is distilled into a symbolic representation communicated to another agent to update its own neural policy. Yet the work cited above has not been applied in A-life.


\section{Proposed Approach}
\label{sec:proposedApproach}

We adopt transformer-based symbolic regression~\citep{biggio2021neural, valipour2021symbolicgpt, kamienny2022end, shojaee2023transformer, lalande2023transformer, vastl2024symformer} and apply it to the problem of distilling neural control policies of A-life agents. While some previous work used manually designed transformer architectures~\citep{biggio2021neural, valipour2021symbolicgpt, kamienny2022end, shojaee2023transformer, lalande2023transformer, vastl2024symformer, wang2026synthesizing} we apply automated neural architecture search (NAS) in a space of transformer architectures. To do so efficiently we combine~\citet{chen2021autoformer}'s autoformers and~\citet{bulitko2022portability}'s triage.


\subsection{Neural Architecture Search}
\label{subsec:neural_architecture_search}

In line with work by~\citet{chen2021autoformer} we first define a space of transformer architectures and initialize the supernet as the largest architecture in that space. Then on each training iteration we randomly select a subnet from the supernet and update its weights. We continue to minimize the training loss of sampled subnets until the validation loss no longer improves over a number of epochs. Once the supernet is trained we run a GA over the space of transformer architectures for a subnet to maximize a fitness function --- the distillation accuracy regularized by the subnet size. 


\subsection{Training the Transformer}
\label{subsec:model_training}

Our transformers receive a set of an agent's percept-action pairs and produce a sequence of tokens representing a symbolic policy. For neural A-life agents percept-action pairs are readily available. However, a symbolic policy for each such neural agent is generally unavailable since the evolution within A-life is entirely neural and does not involve symbolic representations. Thus we do not use data from actual A-life agents to train our transformer. Instead we draw a random symbolic policy $m$ from the given space $M$ of policies and execute $m$ on random agent percepts $p_i$, generating a set of percept-action pairs $\left\{(p_i, m(p_i))\right\}$. The set of pairs forms the transformer input whereas the symbolic policy $m$, expressed as a sequence of tokens in a domain-specific language, constitutes the target output for the transformer. We collate such data from many different policies $m$ to train the transformer.

This method allows us to train the transformer on a large and diverse set of symbolic policies. This is conceptually similar to training large language models (LLM) on text and code from web pages and code repositories and then applying the trained LLM to queries from a specific user. 

To keep our symbolic policies compact and increase their human-readability we define our own A-life-specific language to represent such policies. Doing so renders open-source transformers~\citep{kamienny2022end, shojaee2023transformer} not directly applicable.


\section{Empirical Evaluation}
\label{sec:empiricalEvaluation}

We first present the A-life environment with evolving neural agents. We then define the space of symbolic policies. Finally, we describe the construction and training of the transformer. All parameters in our empirical evaluation were chosen via preliminary experiments to balance the scope of the experiments and the computational cost.


\subsection{A-life}
\label{sec:alife}

We evaluate our transformer-based distillation in an A-life environment --- a variant of the common predator-prey environment~\citep{ackley1991interactions,predation,exag2017,alife2018,exag2018,hastur2018}.

\begin{figure}[htbp]
\centering
\includegraphics[width=0.4\textwidth]{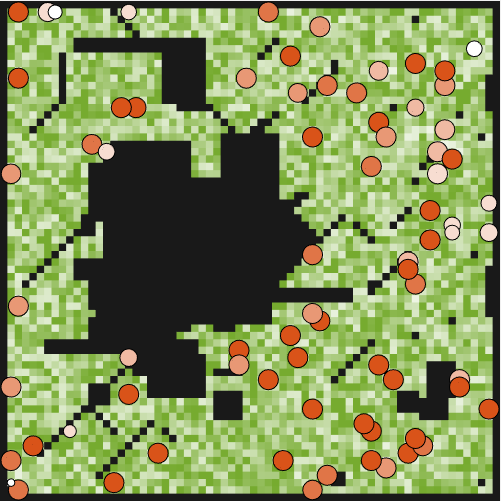}
\caption{The A-life environment.}
\label{fig:alife}
\end{figure}

Our A-life agents exist on a rectangular grid map (Figure~\ref{fig:alife}). Each grid cell is either closed (black squares in the figure) or open. Open cells contain at most one agent (circles in the figure). Agents have five actions available to them: the four moves in the cardinal directions and the action of staying at the current location. Each agent has an energy level. Existing, thinking (i.e., executing the neural control policy) and taking actions all decrease an agent's energy which is replenished by eating grass (shades of green in the figure). When the energy reaches $0$ the agent dies and is removed from the map.
Grass re-grows in each cell up to the maximum value of $1$. The re-growth rate decays over time for a guaranteed eventual population extinction. 

The A-life simulation proceeds in discrete time steps. On each time step the agents (sorted in a random order) perceive the environment around them and run their neural control policies. A policy is a neural network which maps eleven inputs that collectively form the agent's percept to one of the five actions. The eleven inputs are: the grass levels in the four cardinal directions up to a sight radius, inversely scaled by their distance to the agent (inputs 1-4); grass level in the agent's position (input 5); presence of other agents in the four cardinal directions up to a sight radius, inversely scaled by their distance to the agent (inputs 6-9); the agent's energy (input 10) and the agent's normalized age (input 11). All eleven inputs are in the range of $[0,1]$.

While our testbed allows for the neural network architecture to evolve over time, in this paper all agents share the same architecture where the eleven input neurons are connected to eight hidden layers of five neurons each and then connected to the five output neurons. The output neuron with the highest activation indicates the action to take. The network thus has $300$ learnable weights (connection and bias weights). The weights are specified in the agent's genome at birth but can be updated within the agent's lifetime through self-reflection and/or teaching and learning.

Reproduction is asexual. A parental agent has an offspring when it is old enough, has enough energy and the current number of agents does not exceed a pre-specified maximum. The parent gives a part of its energy to the newborn. The offspring's genes are created by copying the weights of the parent and then perturbing them with random values (mutation). Note that similarly to other A-life simulations and unlike the standard GA, our simulation is asynchronous and parents and their children can co-exist. Furthermore, there is no explicit fitness function to sort the population by (as in GA). Finally, the population size is not fixed and can grow up to the maximum size allowed by the specifications. An A-life run ends when all agents die.

\begin{table}[htbp]
\caption{Environment specifications.}
\label{tab:alife}
\centering
{\small \begin{tabular}{r|l}
\toprule
\bf parameter & \bf value\\
\midrule
grass growth & $\nicefrac{1}{33.75}$ \\
grass growth decay & $1-\nicefrac{1}{666.67}$ \\
move energy & $\nicefrac{1}{10}$ \\
live energy & $\nicefrac{1}{8.36}$ \\
power coefficient & $\frac{1}{4.78 \times 10^4}$\\
max grass bite & $\nicefrac{1}{1.33}$ \\
min reproduction energy & $\nicefrac{1}{2.75}$ \\
min reproduction age & $8$ \\
initial agents & $75$ \\
max agents & $61$ \\
mutation rate & $50$ \\
map size & $68 \times 68$ \\
sight radius & $ 7 $ \\
ANN topology & $11 \mhyphen 5 \mhyphen 5 \mhyphen 5 \mhyphen 5 \mhyphen 5 \mhyphen 5 \mhyphen 5 \mhyphen 5 \mhyphen 5$ ($300$ weights) \\
\bottomrule
\end{tabular}}
\end{table}

The environment is specified by a set of parameters (Table~\ref{tab:alife}) as well as the grid map (Figure~\ref{fig:alife}). The parameters and the map were found automatically by stochastically searching the space of parameters and the space of maps via a separate GA. The fitness of a set of environment specifications for this GA was the sum of cubed lifetimes of all agents during an evolution run. Maximizing this fitness function yields environments with numerous yet longer living agents. Since each A-life run is stochastic for a single set of environment specifications we ran $32$ independent A-life runs and took the median value of the number of agents born.


\subsection{Symbolic Policies}
\label{subsec:symbolic_models}

We use an A-life-specific language to define the space $M$ of symbolic policies. Each symbolic policy is a formula defined by the following context-free grammar:

{\small\bea
S &\to& T \midb O_1 \midb O_2 \midb O_3 \midb O_4 \midb O_5 \nonumber \\
T &\to& g_\leftarrow \midb g_\uparrow \midb g_\rightarrow \midb g_\downarrow \midb g_+ \midb a_\leftarrow \midb a_\uparrow \midb a_\rightarrow \midb a_\downarrow \midb e \midb a \midb C \nonumber  \\
O_1 &\to& S+\!+ \midb S-\!- \midb |S| \midb \text{round}(S) \nonumber \\
O_2 &\to& S+S \midb S-S \midb S \times S \midb \frac{S}{S} \midb \nonumber \\
& & \min(S,S) \midb \max(S,S) \midb \argmin(S,S) \midb \argmax(S,S) \nonumber \\
&\vdots& \nonumber \\
O_5 &\to& \min(S,S,S,S,S) \midb \max(S,S,S,S,S) \midb \nonumber \\ & & \argmin(S,S,S,S,S) \midb \argmax(S,S,S,S,S) \nonumber
\eea}

The tokens $g_\leftarrow, g_\uparrow, g_\rightarrow, g_\downarrow, g_+, a_\leftarrow, a_\uparrow, a_\rightarrow, a_\downarrow, e, a$ form the percept of an A-life agent. They represent the grass levels in the four cardinal directions around the agent ($g_\leftarrow, g_\uparrow, g_\rightarrow, g_\downarrow$), the grass at the agent's position ($g_+$), the presence of other agents ($ a_\leftarrow, a_\uparrow, a_\rightarrow, a_\downarrow $), the agent's energy $e$ and the agent's age $a$. Both the grass and the presence of other agents are inversely scaled by the distance.

The production $T$ lists the terminal tokens: the agent's percepts above and the numeric constants: $C = \{0,0.1,\ldots,1\}$. The production $O_1$ lists unary operators (e.g., the absolute value $|S|$). The production $O_2$ lists binary operators and so on until $O_5$, which lists five-argument operators such as $\argmax(S,S,S,S,S)$.

In line with recent work on transformers trained for symbolic regression~\citep{kamienny2022end, shojaee2023transformer, lalande2023transformer, valipour2021symbolicgpt, vastl2024symformer, biggio2021neural} we use the natural sequence-of-tokens notation for our symbolic policies. For instance, the sequence of six tokens 
$\argmax g_\leftarrow \ g_\uparrow \ g_\rightarrow \ g_\downarrow \ g_+ \ \nonumber$ 
represents the formula $\argmax\left(g_\leftarrow,g_\uparrow,g_\rightarrow,g_\downarrow,g_+\right)$
which returns the index of the grid cell around the agent with the highest grass level. We index the agent's actions {\tt left}, {\tt up}, {\tt right}, {\tt down}, {\tt stay} by $1, \ldots, 5$ so the formula above is the symbolic always-move-to-the-highest-grass-neighbor policy.

If on some inputs a formula produces an output outside of the set $\{1,\ldots,5\}$ of action indices the symbolic policy outputs the action index closest to the output. 


\subsection{Training Data for the Transformer}
\label{subsec:data_gen}

To form the training set we generated random sequences of tokens from the grammar. For each sequence, tokens not used in computing the output were removed. If the resulting sequence was empty or exceeded $30$ tokens, it was discarded.

We then formed a set of $1,\!024$ percepts with each of the eleven inputs uniformly randomly drawn from $[0,1]$. If on any of the percepts the formula did not return a numeric value or returned a value outside of the action index range $[1,5]$ the formula was discarded. We additionally discarded formulae whose output was the same for over $99\%$ of the percepts.

Removing duplicate formulae, this process yielded a set of $450,\!100$ unique formulae to train the transformer on. For each formula the transformer input was a $ 64 \times 12 $ matrix of the percept-action pairs. In each row of the matrix the first eleven columns describe a percept and the twelfth column lists the action produced by the formula on the percept. The transformer output was the sequence of tokens representing the formula. We used $ 5 $ random percept-action sets of size $ 64 \times 12 $ each for training. 


\subsection{Defining the Transformer Architecture}

We used the neural architecture search to find an appropriate architecture for the transformer. The space of possible architectures is defined in Table~\ref{tab:nas_search_space}, inspired by existing work~\citep{chen2021autoformer}. The encoder and decoder had the same components with the same space of dimensions.

\begin{table}[htbp]
\caption{Transformer architecture space.}
\label{tab:nas_search_space}
\centering
\begin{tabular}{c|c|c}
\toprule
\bf component & \bf minimum & \bf maximum\\
\midrule
number of layers & $ 1 $ & $ 5 $ \\
number of heads & $ 1 $ & $ 4 $ \\
embedding dimension & $ 64 $ & $ 240 $ \\
Q-K dimension & $ 64 $ & $ 256 $ \\
V dimension & $ 64 $ & $ 256 $ \\
MLP dimension & $ 64 $ & $ 512 $ \\
\midrule
total number of learnables & $ 73,\!581 $ & $ 17,\!296,\!205 $ \\
\bottomrule
\end{tabular}
\end{table}

Percept-action pairs in our training data are permutation invariant. Thus we removed the positional embeddings from the encoder~\citep{kamienny2022end}. To blend the column features we instead added a standard feedforward layer before the encoder~\citep{lalande2023transformer}.

In addition to the previously listed tokens we added three more tokens for transformer training: a beginning-of-sequence token $ \langle bos \rangle $, an end-of-sequence token $ \langle eos \rangle $ and the padding token $ \langle pad \rangle $. During training we added $ \langle bos \rangle $ at the beginning of each formula that served as input for the decoder along with the percept-action pairs for the encoder and we added a $ \langle eos \rangle $ token at the end of each of the target formulae so that the transformer can predict the next token based on the percept-action pairs and previous tokens as context.
 
We used the Adam optimizer~\citep{adam2015method} with a learning rate of $ 10^{-4} $, first and second moment decay constants of $ 0.9 $ and $ 0.999 $ respectively and $\epsilon$ of $ 10^{-8} $. We stopped training when the validation loss did not decrease for over five epochs. We used a mini batch size of $1,\!024$. We randomly partitioned the data into three subsets: $80\%$ for training, $10\%$ for validation during training (to compute the validation loss for early termination) and $10\%$ for testing. 

We brought all formulae to the length of $30$ by adding padding tokens $ \langle pad \rangle $ at the end of shorter formulae. This allowed us to package the formulae into a tensor. When computing gradients during training the padding tokens do not affect gradient computations as they are ignored.

The transformer architecture search started with training the largest transformer (supernet) in our architecture space with $ 17.3 $ million learnable parameters (Table~\ref{tab:nas_search_space}). During the training we did not train the supernet as a whole but rather trained a randomly selected subnet and updated its weights on each iteration~\citep{chen2021autoformer}. We continued the training for $ 90 $ epochs which took $ 14.5 $ hours on a single NVIDIA H100 GPU with 80GB VRAM.

Once the supernet was trained we ran a GA with the population size of $ 20 $ to find a subnet with a high distillation fitness regularized by the number of neurons. The fitness computation was triaged~\citep{bulitko2022portability}: a candidate subnet was first evaluated on $ 2,\!048 $ percept-action pairs. If the candidate was fitter than the previous champion it was subsequently evaluated on $ 10\% $ of the percept-action pairs. Otherwise it was discarded. When the champion fitness did not improve over $ 20 $ generations the GA terminated. 

The GA ran for about $ 1.5 $ minutes and found a subnet (Table~\ref{tab:ga_config}) of $ 5.4 $ million learnable parameters (about $ 32\% $ of the supernet).
The subnet then inherited the weights from the supernet and was trained on the training data set for another $ 69 $ epochs with validation patience of $ 5 $ which took $ 13.5 $ hours.
We found that the accuracy of the evolved subnet is higher when training starts with inherited weights rather than from scratch.

\begin{table}[htbp]
\caption{The NAS-found transformer architecture.}
\label{tab:ga_config}
\centering
{\small\begin{tabular}{c|c}
\toprule
\bf component & \bf maximum size\\
\midrule
encoder depth & $ 5 $ \\
encoder heads & $ (4,\ 1,\ 4,\ 2,\ 1) $ \\
encoder embedding dimension & $ 150 $ \\
encoder Q-K dimension & $ (139,\ 207,\ 85,\ 177,\ 138) $ \\
encoder V dimension & $ (223,\ 99,\ 97,\ 256,\ 183) $ \\
encoder MLP dimension & $ (330,\ 125,\ 459,\ 277,\ 147) $ \\
\midrule
decoder depth & $ 5 $ \\
decoder heads & $ (4,\ 4,\ 1,\ 1,\ 2) $ \\
decoder embedding dimension & $ 205 $ \\
decoder Q-K dimension & $ (209,\ 124,\ 169,\ 169,\ 193) $ \\
decoder V dimension & $ (200,\ 156,\ 148,\ 146,\ 171) $ \\
decoder MLP dimension & $ (333,\ 265,\ 348,\ 240,\ 497) $ \\
\bottomrule
\end{tabular}}
\end{table}


\subsection{Distillation of Neural Agents}
\label{sec:distillationOutsideOfAlife}

We evaluated our trained transformer on distilling A-life's neural policies into symbolic policies and compared the performance to GA~\citep{bulitko2022portability, wang2023synthesizing} and decision trees.

To form the data set we conducted $ 1,\!024 $ A-life runs.
A total of $ 5,\!613,\!559 $ agents lived during those runs. Within each A-life run early agents have random neural policies. Thus to give evolution time to work we discarded all agents who died before time step $105$. To focus on more successful agents we additionally discarded all agents who lived fewer than $105$ time steps or made fewer than $75\%$ energy-gain-maximizing action choices or whose neural policy's most common action on random percepts accounted for at least $90\%$ of actions taken. The filtering process left us with $243$ agents. 

For each of the $243$ agents we ran its neural policy on $1,\!024$ random percepts and recorded the action given by the neural policy. We then drew from the $1,\!024$ percept-action pairs $ 512 $ for distillation and another $ 512 $ for testing the resulting symbolic policies.
We then compared four distillation methods on this data set as follows.

\smallskip{\bf Decision Trees.} We fitted a classification and regression tree (CART)~\citep{breiman1984classification} to the percept-action pairs using MATLAB's built-in \texttt{fitctree} function. The tree partitions the percept space by recursively selecting the split that minimizes the Gini impurity and classifies each leaf region with the plurality action. We evaluated three tree sizes by limiting the maximum number of splits to $ 5, 10$ and $ 20 $. 

\smallskip{\bf GA.} We ran genetic algorithms with a population of $384$ candidate formulae. On each generation the formulae were evaluated for their distillation accuracy on the percept-action data set. The fitness of each formula was its accuracy regularized by the number of tokens in the formula. The fittest $128$ formulae were selected as parents and produced $128$ offspring formulae via crossover and mutation. Each offspring was added to the population twice: as is and with tokens not used in calculation of the formula value removed. A running fitness champion was maintained. If the fitness did not improve over $100$ generations the population was reinitialized. The process terminated when the time limit was reached and the fitness champion was returned as the result of the distillation. We used six different time limits: $ 5, 10, 30 $ seconds and $ 1, 3, 45 $ minutes.

\smallskip{\bf Portfolio GA.} Given the stochastic nature of the GA we also ran a portfolio GA with a portfolio of size four. Each member of the portfolio was the previously described GA but run with a different random number generator seed. When all four GA runs finished we selected the formula with the highest distillation fitness. 

\smallskip{\bf Transformer.} When using the transformer for distillation besides the percept-action pairs we gave it a starting token $ \langle bos \rangle $ as the context. When given a context sequence of any length the transformer returned a sequence of the same length with the last token being the newly predicted token. Each new token was appended to the previous context sequence for the transformer to predict the next token. The process proceeded until either the transformer predicted an $ \langle eos \rangle $ token or the length of the sequence reached its maximum length. During distillation we randomly sampled $ 64 $ from the given $ 512 $ percept-action pairs as the input for the transformer to predict a formula, similar to the bagging trick described in existing work~\citep{kamienny2022end, shojaee2023transformer, wang2026synthesizing}. We applied beam search with width $ 1 $ and $ 5 $~\citep{vaswani2017attention} to investigate the effect of beam width on accuracy.

\begin{figure}[htbp]
\centering
\includegraphics[width=0.7\textwidth]{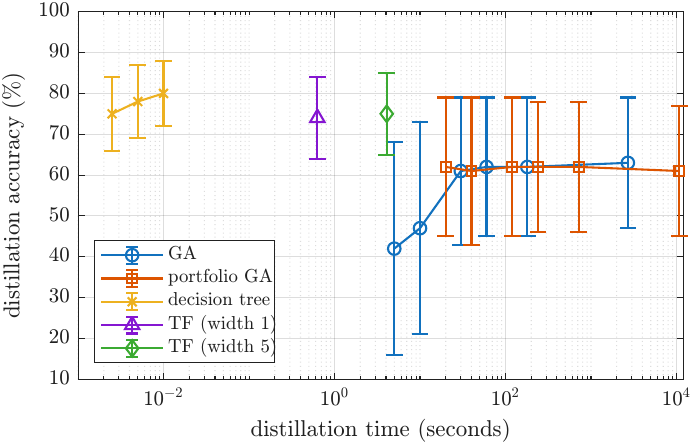}
\caption{Distillation of percept-action pairs of A-life agents.}
\label{fig:distill}
\end{figure}

\smallskip{\bf Observations.} The transformer with beam search with beam widths of $ 1 $ and $ 5 $ had the average distillation times of $ 0.63 $ and $ 4.08 $ seconds and their average distillation accuracy on the test set surpassed any GA including that with the $ 45 $-minute time limit (Figure~\ref{fig:distill}).

\smallskip{\bf Human readability of symbolic policies.} Both the GA and the transformer tend to produce symbolic policies of similar length. We will now discuss their human readability on a specific example from a preliminary experiment. Consider an evolved A-life agent that lived for $434$ time steps and died at time step $890$.
Its actions on random percepts were diverse with the most common action constituting $ 27\% $ of all actions.

When run on a set of random percepts and the corresponding actions of the agent's neural policy, the transformer output the symbolic policy $\argmax(g_\leftarrow,g_\uparrow,g_\rightarrow,a)$ which is readily understood as go left/up/right when that's the direction with the most grass and the agent's normalized age is below that grass amount. Go down when the agent's normalized age becomes higher than the three grass amounts. This symbolic policy has the accuracy of about $99\%$ on the set of $512$ random percepts. In fact only three of the agent's $512$ actions are mislabeled by the symbolic policy: two {\tt right} actions are mislabeled as {\tt left} and one {\tt right} is mislabeled as {\tt up}.
 
While a decision tree can be human-readable~\citep{rudin2019stop} even a shallow tree with $ 5 $ splits produces a branching structure that must be traced path by path to understand the policy. In practice this makes decision trees more difficult for a human to parse than compact formula-based symbolic policies produced by the GA or the transformer.


\subsection{Policy Distillation within A-life}
\label{sec:alifeDistillation}

When distillation of agent neural policies takes place during an agent's lifetime within A-life it can affect both individual agents and the population as a whole. In this section we consider two ways of using distillation within A-life.

\smallskip{\bf Self-reflection.} Every $25$ time steps of its life each A-life agent ran the distillation process if its energy was at least $75\%$ of the maximum. To do so, the agent ran its neural policy on $256$ random percepts and recorded the actions returned by the policy. It then invoked the transformer to produce a symbolic policy approximating the percept-action set and, therefore, its neural policy. The resulting symbolic policy was then used to generate $32,\!768$ percept-action pairs. These were in turn used to update the weights of its own neural policy in an act of self-reflection. While perfectly accurate distillation would likely result in minimal changes to the neural weights, a symbolic bias in the distillation can affect the neural weights severely.

\smallskip{\bf Teaching and learning.} An agent then engaged in teaching by distilling its neural policy and sending the resulting formula to a random agent in the population. The receiving agent updated its own neural policy in the manner described above. This mechanism allows the agents to share their control knowledge in a compact fashion as transmitting a short formula representing the symbolic policy is faster and easier than transmitting percept-action pairs or neural weights.

\smallskip{\bf Symbolic subspaces.} Self-reflection, teaching and learning pass an agent's neural knowledge through the symbolic space. In our A-life environment this is the only mechanism by which the agents learn within their lifetime. Therefore the learning bias imposed by the symbolic space can in principle affect agents individually and their population as a whole.

\begin{figure}[htbp]
\centering
\includegraphics[width=0.85\textwidth]{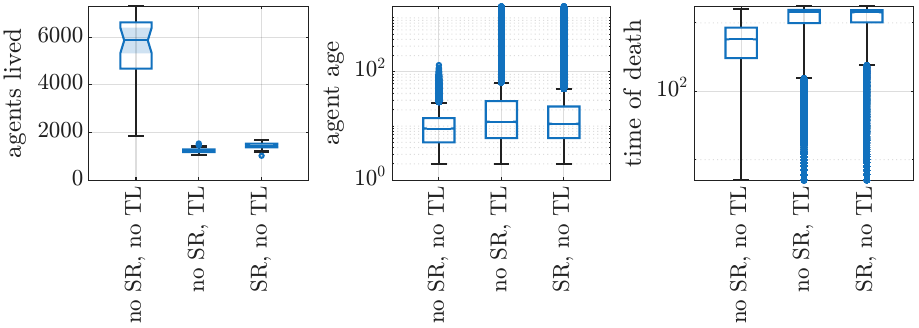}
\caption{Influence of self-reflection (SR) and teaching and learning (TL) on A-life population dynamics with a strong symbolic bias.}
\label{fig:argmaxG}
\end{figure}

In a preliminary experiment with a smaller A-life environment we first considered a trivial space of symbolic policies containing only a single formula $\argmax\left(g_\leftarrow,g_\uparrow,g_\rightarrow,g_\downarrow,g_+\right)$ which always takes the action in the direction of the most grass. Having only a single symbolic policy imposes a strong learning bias on self-reflection and teaching and learning. The results from $32$ A-life runs suggest that such biased self-reflection and teaching and learning make the agents live longer and die later (Figure~\ref{fig:argmaxG}). 

\begin{figure}[htbp]
\centering
\includegraphics[width=0.6\textwidth]{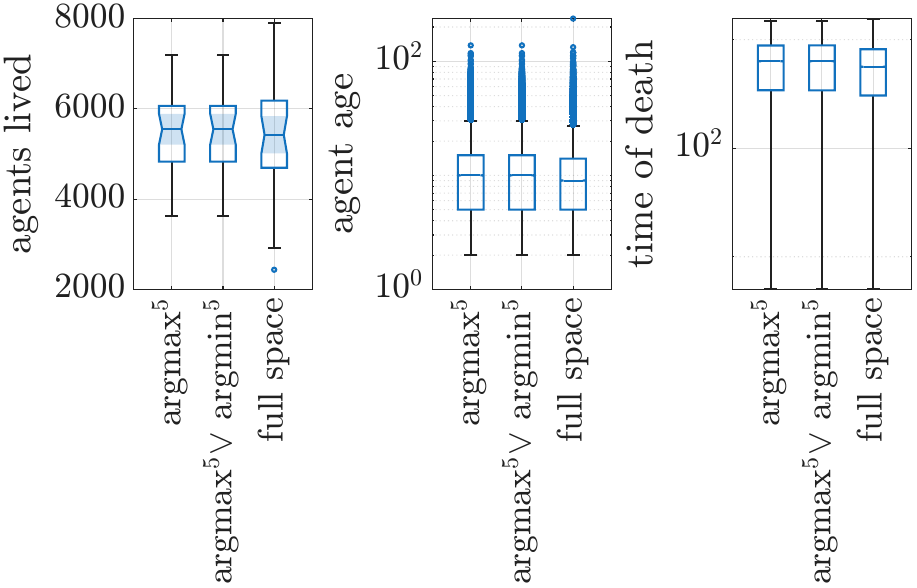}
\caption{Different subspaces of symbolic policies. Both self-reflection and teaching and learning are performed.}
\label{fig:subspaces}
\end{figure}

We then used the larger A-life environment described earlier in the paper. We incrementally weakened the learning bias by allowing the transformer to use progressively larger subspaces of the full space of symbolic policies. The subspaces were: $\argmax\nolimits^5$ which is the space of all formulae starting with a $5$-argument $\argmax$; $\argmax\nolimits^5 \vee \argmin\nolimits^5$ which is the space of all formulae starting with a $5$-argument $\argmax$ or $\argmin$; and the full space of all formulae in our grammar. These subspaces are strict subsets of each other: $\argmax\nolimits^5 \subset \argmax\nolimits^5 \vee \argmin\nolimits^5 \subset \text{full space}$. All formulae were restricted to having at most $ 30 $ tokens.

\begin{figure}[htbp]
\centering
\includegraphics[width=0.6\textwidth]{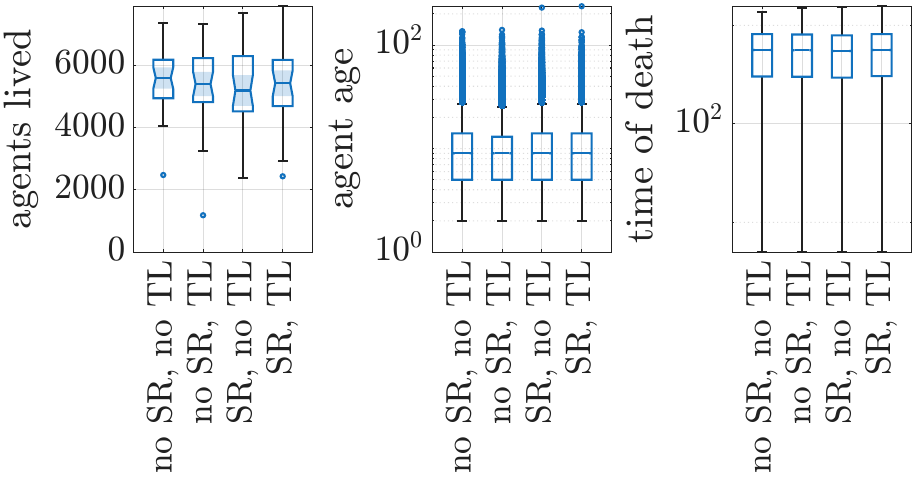}
\caption{Influence of self-reflection (SR) and teaching and learning (TL) on A-life population dynamics with the transformer and the full symbolic space.}
\label{fig:tf_no_bias}
\end{figure}

With the smaller and simpler A-life environment and the high symbolic bias, combining teaching and learning with self-reflection affects the medians for the number of agents lived and the time of death (Figure~\ref{fig:argmaxG}). However, removing the symbolic bias and using a larger, more complex A-life environment eliminates the effects (Figure~\ref{fig:tf_no_bias}).

\section{Shortcomings \& Future Work}
\label{sec:futureWork}

Currently the effects of self-reflection and teaching and learning are specific to the A-life environment with only the smaller, simpler A-life environment and stronger symbolic bias showing significant effects. Future work will investigate why the same trends are not observed with the larger, more complex A-life environment and larger symbolic policy spaces.

Despite the orders-of-magnitude reduction in run time relative to the GA, running the transformer with $ 5.4 $ million weights is still notably more expensive than running the agent's own $ 300 $-weight neural policy. Had the distillation not been free the A-life agents would run out of energy and die after a single distillation. Thus, it will be of interest to evolve distillers together with the agent control policy within A-life evolution and see if smaller, more practical distillers emerge similar to previous work in reinforcement learning~\citep{landajuela2021discovering}. Furthermore, our agents currently self-reflect and teach involuntarily. Future work will explore environments which are conducive to emergence of self-reflection and teaching and learning as a voluntary action taken by agents.

Some of our preliminary results showed that the space of symbolic policies can indeed affect population dynamics. Thus, it will be of interest to evolve symbolic policy spaces in A-life alongside the agents and their distillers. An interesting question is whether, for any agent, there necessarily exists a task that the agent can perform but cannot symbolically articulate (distill) due to the computational overhead of such distillation. In particular if an articulation of a policy is a computational shortcut to the original policy then computational irreducibility may restrict the type of policies that can be articulated~\citep{nks,wolfram2024,berkovich2025lifegpt}. Consequently, learning agents in A-life may have to use their own life experience to complement the symbolic policy they receive from a teaching agent.


\section{Conclusions}
\label{sec:conclusions}

Neural control policies of A-life agents are generally opaque to an external observer. Approximating them with compact symbolic policies can help human understanding. We proposed to use a transformer to distill neural policies of autonomous agents in A-life. Once trained, the transformer surpasses the distillation accuracy of genetic algorithms while being substantially faster. The fast distillation afforded by the transformer makes it more practical to deploy it on the agents themselves which can enable practical teaching and learning within a multi-agent system. 


\section*{Acknowledgments}
We thank the Natural Sciences and Engineering Research Council of Canada (NSERC) for the funding and the Digital Research Alliance of Canada (DRAC) for the computational resources.


\bibliographystyle{plainnat}
\bibliography{refs}

\end{document}